\documentclass[12pt]{article}

\usepackage{arxiv}

\usepackage{graphicx} % Required for inserting images
\usepackage{amsmath, amsfonts, amssymb, amsthm}
\usepackage{hyperref}

\newcommand\cC{\mathcal C}

\usepackage[math]{cellspace}
\usepackage{xcolor}
\definecolor{purple}{rgb}{0.5, 0.1, 0.9}

\title{Leveraging Non-linear Dimension Reduction and Random Walk Co-occurrence for Node Embedding}
\author{Ryan DeWolfe
    \thanks{Department of Mathematics, Toronto Metropolitan University, Toronto, Canada. ryan.dewolfe@torontomu.ca}
}
\date{}

\begin{document}

\maketitle

\begin{abstract}
    Leveraging non-linear dimension reduction techniques, we remove the low dimension constraint from node embedding and propose COVE, an explainable high dimensional embedding that, when reduced to low dimension with UMAP, slightly increases performance on clustering and link prediction tasks.
    The embedding is inspired by neural embedding methods that use co-occurrence on a random walk as an indication of similarity, and is closely related to a diffusion process.
    Extending on recent community detection benchmarks, we find that a COVE UMAP HDBSCAN pipeline performs similarly to the popular Louvain algorithm.
\end{abstract}
\keywords{Node Embedding, Random Walks, Dimension Reduction, Community Detection}

\section{Introduction}
Unsupervised node embedding algorithms~\cite{embedding_review,embedding_review2} assign each node of a graph to a low dimension vector, and allows graph mining~\cite{mining_complex_networks} to leverage existing data science tools for tasks like visualization, clustering (also called community detection)~\cite{fortunato_review}, and link prediction~\cite{linkpred}.
Some existing methods, like DeepWalk~\cite{deepwalk} and node2ec~\cite{node2vec}, combine random walks with feature learning techniques that were originally developed for natural language processing~\cite{skipgram,negative_sampling} to efficiently learn low dimensional embedding vectors.
The key assumption in these algorithms is that nodes that are often nearby in the random walk should appear close in the embedding space.

\begin{figure}
    \center
    \includegraphics[width=\linewidth]{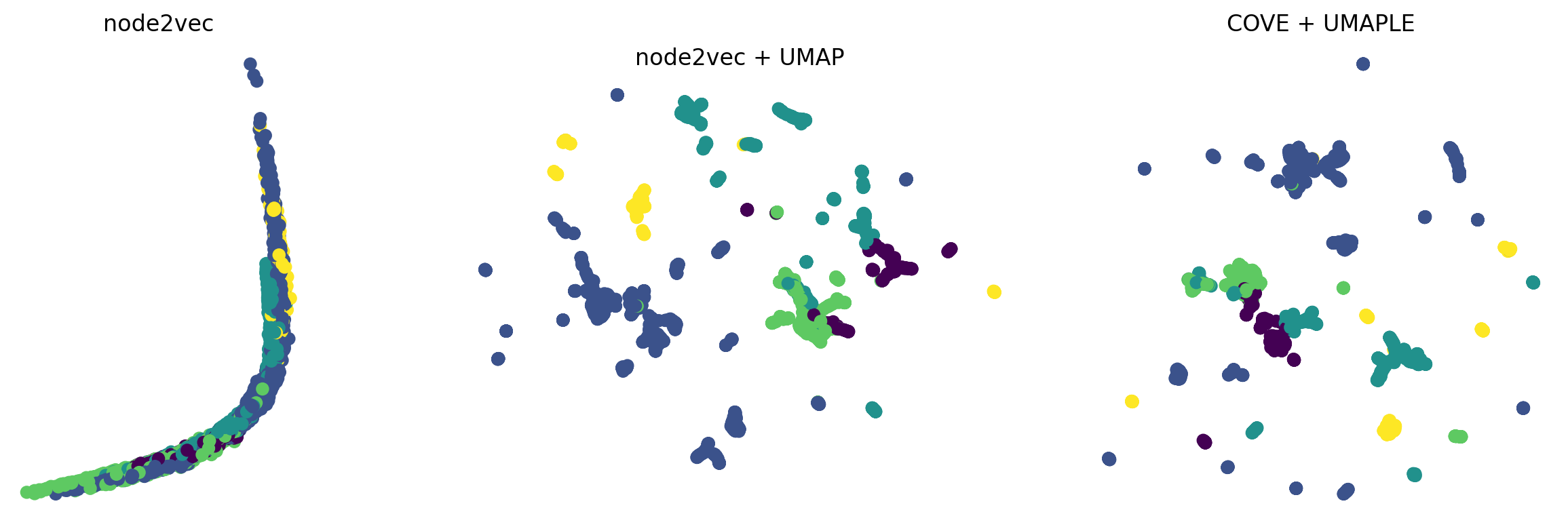}
    \caption{
        Example of embeddings produced by node2vec, node2vec+UMAP, and the proposed COVE+UMAPLE algorithm.
        The graph is the world-wide airport graph of airports and their connections via direct flights, with ground truth communities (colored) corresponding to the continents.
        Using node2vec directly to 2 dimensions does not separate any clusters, whereas using node2vec then UMAP or COVE then UMAPLE does (with similar visual quality).
    }
    \label{fig:example}
\end{figure}

Unfortunately, due to the structure of the representation learning methods, directly embedding to a very low dimension (often 2d for visualization) does not preserve meso-scale structures like communities (see Figure~\ref{fig:example}).
Luckily, we can first embed to a moderate dimension (128 for example) and use a dimension reduction technique like UMAP~\cite{umap} or t-SNE~\cite{tsne} to better preserve communities in the low dimension space.

However, the inputs to the dimension reduction techniques do not need to be low dimensional.
In fact, node embeddings need not be low-dimensional either, and we suspect the low-dimension constraint is due to the curse of dimensionality that makes high-dimensional embeddings difficult for the existing data science methods.
By removing the low-dimension constraint from the embedding step, we can propose high-dimensional vectors and leverage modern dimension reduction techniques to later decrease the dimension.
In this paper, we propose COVE, an explainable high dimensional embedding method inspired by random walks co-occurrence and closely linked to diffusion processes, that, when combined with non-linear dimension reduction techniques, leads to slightly improved low dimensional embeddings (as measured by an unsupervised score~\cite{cge} and performance on clustering and link prediction tasks).
We also expand on previous studies that test embeddings for clustering tasks~\cite{clustering_via_embeddings,neural_embeddings} by replacing K-means~\cite{kmeans} with HDBSCAN~\cite{hdbscan}, and find that a pipeline of COVE, UMAP, and HDBSCAN performs similarly to the very popular Louvain~\cite{louvain} community detection algorithm but still worse than the state-of-the-art ECG~\cite{ecg}.
Thus, we conclude that leveraging dimension reduction techniques allows for more explainable embeddings and a slight performance improvement.

\section{Review of neural embedding techniques}

\subsection{Neural Representation Learning Models}\label{sec:ne}
Underlying neural embedding techniques are methods borrowed from natural language processing that were developed to create word vector representations from text data.
These models take a set of sequences of tokens (words) and produce a $d$ dimensional representation vector for each token.

More formally, let $W^{(i)} = w^{(i)}_1, w^{(i)}_2, \dots w^{(i)}_{n_i}$ be a sequence of tokens, $W = \{W^{(i)}: i \in \mathcal{I}\}$ for a set of documents indexed by $\mathcal{I}$, and define $\mathcal{W}$ as the set of tokens appearing at least once.
Furthermore, define parameters $d \in \mathbb{N}$ (dimension) and $L \in \mathbb{N}$ (context radius).
For each token $w^{(i)}_t$, its context $C(w^{(i)}_t)$ is the multiset of tokens $w^{(i)}_{t-L}, w^{(i)}_{t-L+1}, \dots w^{(i)}_{t-1}, w^{(i)}_{t+1}, \dots w^{(i)}_{t+L-1}, w^{(i)}_{t+L}$, subject to the indices existing due to sequence boundaries.
Finally, each token $w$ is given a context (input) vector $c_w \in \mathbb{R}^d$ and a representation (output) vector $v_w \in \mathbb{R}^d$ that are initially random and will be learned during the training process.

Mikolov et al.~\cite{skipgram} propose two architectures for learning the context and representation vectors.
First, the Continuous Bag of Words (CBOW) model uses the average context vector to predict the representation, maximizing the log-loss function
\begin{equation}\label{eq:cbow}   
    \sum_{i \in \mathcal{I}} \sum_{t=1}^{n_i} \log P \left( w^{(i)}_t | \overline{C(w^{(i)}_t)} \right).
\end{equation}
Next, the better performing and more popular Skip-gram model reverses the process and uses each context vector predict the token's representation. The log-loss function is
\begin{equation}\label{eq:skipgram}
    \sum_{i \in \mathcal{I}} \sum_{t=1}^{n_i} \sum_{w \in C(w^{(i)}_t)} \log P(w|c_{w^{(i)}_t}).
\end{equation}
Both models would define $P(w|c)$ using a softmax,
\begin{equation}\label{eq:softmax}
    P(w|c) = \frac{\mathrm{exp}(v_w \cdot c)}{\sum_{t \in \mathcal{W}} \mathrm{exp}(v_w \cdot c_t)},
\end{equation}
but the denominator makes this approach computationally inefficient.
The models are proposed using a hierarchical softmax~\cite{hierarchical_softmax} to approximate equation~\ref{eq:softmax}, but quickly switch to a more efficient negative sampling method~\cite{negative_sampling}.
Negative sampling is a variation of noise contrastive estimation that assumes a context vector should have a low probability of predicting a random token (with the assumption they are unlikely to occur together, hence a negative sample).
The proposed loss function uses the sigmoid of the dot product as a score, and each training example (w,c) randomly chooses $k$ representation vectors to ``push away'' from $c$.
Let $R$ be a discrete random variable over $\mathcal{W}$ with $P(R=w)$ proportional to total number of occurrences of $w$ in $W$ (or a similar distribution that down-weights very common tokens~\cite{negative_sampling}).
Negative sampling can be used with either the CBOW or SkipGram models by using the substitution
\begin{equation}
    \log P(w|c) = \log \sigma({v_{w}} \cdot c) + \sum_{i=1}^k \log \sigma(-v_{w_i \sim R} \cdot c)
\end{equation}
in the loss functions defined in equation \ref{eq:cbow} or \ref{eq:skipgram} respectively.

Levy and Goldberg~\cite{skipgram_factorization} find that SkipGram with Negative Sampling (SGNS) is implicitly factoring a matrix.
Let $n = |\mathcal{W}|$ and fix an ordering $w_1, w_2, \dots w_n$ of $\mathcal{W}$.
Let $V \in \mathbb{R}^{n \times d}$ be the representation matrix, created by stacking the representation vectors such that the $i^{th}$ row corresponds to $v_{w_i}$.
Similarly define the context matrix $C \in \mathbb{R}^{n \times d}$.
SGNS learns $V$ and $C$ such that $VC^{\top} \approx M \in \mathbb{R}^{n \times n}$, the shifted pointwise mutual information matrix.
Let $\#_v(w)$ be the number of times $w$ appears in $W$, and let $\#_c(w)$ be the number of times $w$ appears as context, $\#_c(w) = \sum_{i \in \mathcal{I}} \sum_{w' \in W^{(i)}} \chi(w \in C(w'))$.
Also define $\#(w_i, w_j)$ as the number of times $w_j$ is in the context of $w_i$, $\#(w_i, w_j) = \sum_{i \in \mathcal{I}} \sum_{w' \in W^{(i)}} \chi(w' = w_i) (w_j \in C(w'))$.
The entries of $M$ are computed as
\begin{equation}
    M_{ij} = \mathrm{PMI}(w_i, w_j) - \log k = \log \left(\frac{\#(w_i, w_j) \sum_{w' \in \mathcal{W}} \#_c(w')}{\#_v(w_i) \cdot \#_c(w_j)}\right) - \log k.
\end{equation}
The factor $\sum_{w' \in \mathcal{W}} \#_c(w')$ in the numerator comes from the definition of PMI using probabilities, which normalizes each $\#$ term by the total number of context tokens.

They propose computing this matrix directly, or alternatively a sparse variant $M'$ called the shifted positive PMI, where
\begin{equation}
    M'_{ij} = \max\{\mathrm{PMI}(w_i, w_j) - \log k, 0\},
\end{equation}
and factoring directly with SVD.

\subsection{Learning Node Representations}
Due to the success of the neural methods, several techniques were developed to apply the representation learning to graph embedding~\cite{deepwalk,node2vec}.
Each technique defines a method for generating a set of node sequences using random walks, and then learns node representations using a neural model (usually SGNS).

The first method is DeepWalk~\cite{deepwalk}, which uses a standard walk.
For a graph $G$ with nodes $V$ and edges $E$, a random walk of length $\ell$ starting from node $v_1 \in V$ is a random sequence $v_1, v_2, \dots v_\ell$ where
\begin{equation}
    P(v_{i+1} = u|v_i) = \frac{\chi(uv_i \in E)}{deg(v_i)}.
\end{equation}
Or equivalently, the random walk will transition from node $v$ to one of its neighbours, each with probability $1 / deg(v)$ (for weighted graphs the transition probability is proportional to edge weights).
DeepWalk creates its corpus $W$ by sampling $\gamma$ random walks of length $\ell$ starting from each node $v$; both $\gamma$ and $\ell$ are user specified parameters.

In a generalization of DeepWalk, node2vec~\cite{node2vec} proposes adding parameters to bias the random walk so that different search strategies can be employed.
A random walk in node2vec has parameters $p$ and $q$ in $(0,\infty)$ to control, respectively, the probability of backtracking and moving away from the previous node.
For a random walk of length $\ell$ starting from node $v_1$, the second node $v_2$ is sampled as before, and then
\begin{equation}
    P(v_{i+1} = u|v_i, v_{i-1}) \propto \begin{cases}
        \frac{1}{p} & \text{if } dist(u, v_{i-1}) = 0 \\
        1 & \text{if } dist(u, v_{i-1}) = 1 \\
        \frac{1}{q} & \text{if } dist(u, v_{i-1}) = 2
    \end{cases}
\end{equation}
measuring distance as the length of the shortest path.
The corpus is generated with the same process as in DeepWalk, and setting $p = q = 1$ makes node2vec equivalent to DeepWalk.
However, beyond intuitive explanations about the behaviour of the random walks, there is no unsupervised method or general guidance for selecting parameters $p$ and $q$, and they are often left as $1$ by default~\cite{pecanpy,clustering_via_embeddings,neural_embeddings}.

A third embedding technique often discussed as an application of neural methods is LINE~\cite{line}, but as it does not use a random walk, instead opting to define all token-context pairs explicitly, we will omit discussion for the sake of brevity.

Following the results of Levy and Goldberg~\cite{skipgram_factorization}, Qiu et al.~\cite{neural_embedding_as_matrix} note that the use of SGNS in these graph embedding methods means they are also implicitly factoring matrices and find have connections to spectral techniques~\cite{chung_spectral}.
They further suggest computing and factoring the DeepWalk matrix directly in a new method called NetMF.

\section{Method}
The key idea from the reviewed work is leveraging close co-occurrence of nodes in a random walk to create an embedding.
Our proposed embedding for each node $v$ can be simply described as the distribution of close co-occurrences with $v$ in a random walk.
If we consider a standard walk, like the one used in DeepWalk, we can explicitly compute these vectors for a given context window size $L$ by multiplying transition matrices.
Let $\hat{A}$ be the transition matrix for a standard random walk, equivalent to the row-normalized adjacency matrix of $G$ (fixing an arbitrary ordering of the nodes $v_1, v_2, \dots v_n$).
The probability of randomly walking from $v$ to $u$ in $i$ steps can be computed as the entry $(\hat{A}^i)_{uv}$.
Weighting each distance of co-occurrence equally (some packages allow for a kernel on distance from the center of the context, but give little guidance on how to set it) we can compute the probability of node $v$ occurring after node $u$ as the $T_{uv}$ in the matrix
\begin{equation}
    T = \sum_{i=1}^L \hat{A}^{i}.
\end{equation}
And to allow for co-occurrence in either direction, compute
\begin{equation}
    \psi = T + T^\top
\end{equation}
and call $\hat{\psi}$ the row normalization of $\psi$.
The row $\hat{\psi}_i$ is the proposed embedding vector for node $i$.

Note that our definition is symmetrized truncated diffusion process.
Diffusion processes, such as personalized page rank~\cite{ppr}, heat kernel pagerank~\cite{hk} or Katz centrality~\cite{katz}, take the form
\begin{equation}
    \sum_{i=1}^\infty \theta_i \hat{A}^i
\end{equation}
with coefficients $\theta_i$ determining the specific process.
Thus, there is an interpretable meaning to using a window kernel to weight co-occurrences based on how close they are in the random walk.

If $G$ is large or not extremely sparse, computing powers of $\hat{A}$ becomes difficult.
Additionally, although probably feasible, a closed form of the co-occurrence probabilities for a biased random walk would be complicated~\cite{neural_embedding_as_matrix}.
Thus, we suggest approximating $\hat{\psi}$ by sampling random walks analogous to the reviewed neural embedding methods.
Using either a standard or biased random walk, sample a training corpus $W$ by starting $\gamma$ random walks of length $\ell$ from each node.
The approximate matrix $\tilde{\psi}$ is then computed as the row-normalization of $\#$ (as defined in Section \ref{sec:ne}).

While the definition of $\psi$ is intuitive and $\hat{\psi}$ is a computationally efficient approximation, these representation vectors are high dimensional.
Using the Hellinger distance between distributions (a scaled euclidean distance between vectors), we can apply a dimension reduction method to try and maintain distances but in a low dimensional space.
Truncated SVD~\cite{svd} is an extremely fast linear method that will attempt to maintain all distances.
Alternatively, there has been significant progress in non-linear dimension reduction techniques~\cite{tsne,umap,localmap} that focus on local distances.
To limit the scope of this paper, our non-linear dimension reduction technique will be UMAP~\cite{umap} due to its popularity, speed, and open-source python implementation~\cite{umap_code}.
As we will see in the upcoming experiments, the choice of dimension reduction technique has a considerable impact on the performance of the embedding for downstream tasks, and separating the embedding method from the dimension reduction allows for different decisions depending on the specific task.

During the experiments, many of the UMAP runs reported a warning that the spectral initialization~\cite{laplacian_eigenmaps} of the low dimensional vectors failed, and that it automatically fell back to a random initialization.
It has been noted that the initialization of UMAP is important~\cite{umap_init1,umap_init2}, and that spectral initialization outperforms a random initialization.
Luckily, we can use the graph itself to inform the initialization step.
We test initialization via a spectral embedding~\cite{chung_spectral} of the graph $G$, which is the same method that Laplacian Eigenmaps applies to UMAP's internal k-nearest-neighbors graph for initialization.
We refer to this method as UMAPLE in the following sections.

\section{Experiments}
For the following experiments, we use the Pecanpy~\cite{pecanpy} implementation of node2vec with parameters $p=q=1$.
The number of walks per node is fixed to 10, the length of each walk is fixed to 40, and the width of the window is set to 13 (6 on each side of the center).
For the node2vec+UMAP methods, node2vec embeds into 128 dimensions and then UMAP is used to reduce to the stated dimension (possibly also 128).
We use the open source python implementations of UMAP~\cite{umap_code} and HDBSCAN~\cite{hdbscan_code}, and the Scikit-learn~\cite{scikitlearn} implementations of TruncatedSVD and K-means.

\subsection{Data} \label{sec:data}
We consider both real and synthetic datasets that have ground truth communities.
We gather a variety of real graph datasets with ground truth communities which are summarized in Table \ref{ref:tab_data}. Any ground truth clusters of size less than or equal to 5 were removed and set as outliers (only eu-core and as were effected).
\begin{table}
    \centering
    \begin{tabular}{|Sc|Sc|Sc|Sc|Sc|Sc|Sc|Sc|Sc|Sc|} \hline
    Name & Ref & URL & Domain &  $|V|$ & $|E|$ & $|\mathcal{C}|$ & $s_0$ & $\eta$ & $\xi$\\
    \hline
    Football & \cite{football,football_coms} & \cite{com_graphs} & Competition & 115 & 613 & 10 & 12 & 1.00 & 0.38 \\ 
    Primary1 & \cite{primary} & \cite{com_graphs} & Contact & 236 & 5899 & 11 & 0 & 1.00 & 0.61 \\ 
    Primary2 & \cite{primary} & \cite{com_graphs} & Contact & 238 & 5539 & 11 & 0 & 1.00 & 0.58 \\ 
    Eu-core & \cite{eucore} & \cite{com_graphs} & Email & 1005 & 16385 & 35 & 18 & 1.00 & 0.68 \\
    Eurosis & \cite{eurosis} & \cite{com_graphs} & Web & 1285 & 6462 & 13 & 0 & 1.00 & 0.18 \\ 
    Cora  & \cite{cora} & \cite{com_graphs} & Citation & 2708 & 5278 & 7 & 0 & 1.00 & 0.19 \\ 
    Airport & \cite{airports} & \cite{konect} & Transport & 2898 & 15564 & 5 & 0 & 1.00 & 0.16 \\ 
    Blogcatalog & \cite{blogcatalog} & \cite{asu_graphs} & Social & 10312 & 333983 & 39 & 0 & 1.40 & 0.81 \\ 
    Cora Large & \cite{cora_large} & \cite{com_graphs} & Citation & 23166 & 89157 & 70 & 0 & 1.00 & 0.46 \\ 
    As & \cite{as} & \cite{com_graphs} & Web & 23752 & 58416 & 96 & 165 & 1.00 & 0.56 \\
    \hline
    \end{tabular}
    \vspace{1em}
    \caption{
        Table of empirical values for real world datasets with ground truth communities.
        The columns $|\mathcal{C}|$ denotes the number of communities, $s_0$ the number of outliers, and $\eta$ the average number of communities per non-outlier node.
        The proportion of inter-community edges $\xi$ is (approximately) analogous to the $\xi$ parameter in the ABCD model.
    }
    \label{ref:tab_data}
\end{table}
For synthetic graphs with ground truth communities, we use the \textit{Artificial Benchmark for Community Detection} (ABCD)~\cite{abcd}.
The ABCD model similar to the popular LFR~\cite{lfr} model, and feature power-law degrees and community sizes.
The main difference is that the ABCD noise parameter $\xi$ allows for a smooth transition from disjoint communities to a completely random graph.
The degrees and community sizes are sampled from truncated power law distributions with minimum, maximum, and exponent values of 3, 70, 2.5 and 15, 700, 1.5 respectively.
The ABCD model randomly generates a community graph for each community and a global background graph containing all the noise edges using the configuration model.
The noise parameter $\xi \in [0,1]$ controls the proportion of the degree of each node that is in the background graph, and so when $\xi = 1$ the graph is completely random.

\subsection{Unsupervised Evaluation}
To address the variety and stochastic nature of graph embedding algorithms, Kami\'{n}ski et al.~\cite{cge} developed a pair of divergence scores to measure the quality of an embedding.
The first divergence captures the global structure; it uses the Jensen-Shannon divergence to compares the proportions of between within and between communities (detected with ECG~\cite{ecg}) to the expected values under a geometric Chung-Lu null model.
In the geometric Chung-Lu mode, nodes $u$ and $v$ are connected independently at random with probability $p(u,v) \propto w_uw_v(d_{max} - d(u,v))^\alpha$, where $d(u,v)$ is the distance between u and v, $d_{max}$ is the largest distance between nodes in the embeddings, $\alpha$ is a learned parameter, and $w_u, w_v$ are weights chosen so that the expected degree of each node equals its degree in the original graph.
The second score is local, and uses $1 - $AUROC to measure $p(u,v)$ as a predictor of an edge.

\begin{figure}
    \center
    \includegraphics[width=\linewidth]{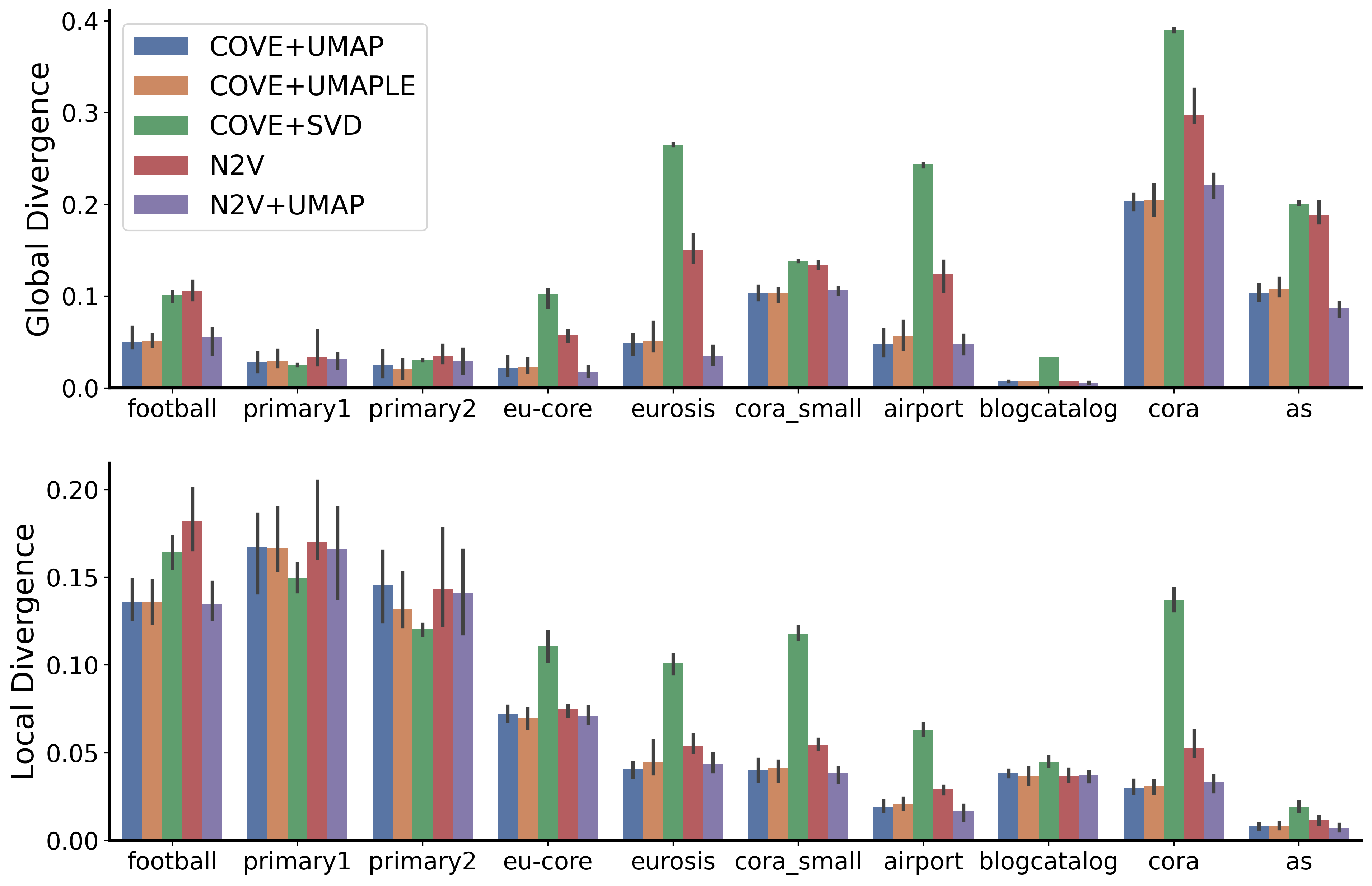}
    \caption{
        Evaluating the quality of each embedding method on real networks using an unsupervised framework~\cite{cge}.
        Each bar represents the average of 10 independent embeddings, with the small bar covering the full range of values, for both the global (top) and local (bottom) divergences are reported.
        In both graphs, a lower score indicates better performance.
    }
    \label{fig:cge}
\end{figure}

In Figure \ref{fig:cge}, we use this framework to evaluate 2 dimension embeddings produced by COVE with UMAP, UMAPLE, and SVD or node2vec either directly to 2d or to 128d and then reduced to 2d with UMAP.
None of the tested methods is clearly out performing the others, although the methods using UMAP slightly out perform the others on the football, cora\_small, and airport graphs.

\subsection{Clustering}\label{sec:clu}
Previous studies~\cite{clustering_via_embeddings,neural_embeddings} have used the K-means~\cite{kmeans} algorithm for clustering the embedding vectors, but have noted that poor performance may be due issues with K-means; for example, K-means struggles with heterogeneous cluster sizes, such as those following a power-law distribution which occur in many real world networks~\cite{clustersizes}.
We also test HDBSCAN~\cite{hdbscan,hdbscan_code}, a density based clustering algorithm that detects the connected components of high density and allows for outliers in low density regions between clusters.
Another benefit of HDBSCAN is that it's only parameter is the minimum cluster size (by default set to 15), which is equally interpretable but less restrictive than the number of clusters parameter required by K-means.

To evaluate the performance of the clustering algorithms, we use a recently developed extrinsic similarity measure \cite{f*} that can handle outliers and overlapping communities.
Extrinsic measures evaluate the ability of a clustering algorithm to produce a specific clustering (usually called the ground truth) that is assumed to be desirable.
Define a clustering $\mathcal{C}$ as a set of non-empty subsets of nodes $V$, and consider two clusterings $\mathcal{C}_1$ and $\mathcal{C}_2$.
To compare two clusters, $C_i$ and $C_j$, we use the $F^*$ score (or Jaccard Index), defined as $F^*(C_i, C_j) = |C_i \cap C_j| / |C_i \cup C_j|$.
However, there is not a matching of clusters from $\mathcal{C}_1$ to $\mathcal{C}_2$, so we take the maximum similarity when compared to each cluster in $\mathcal{C}_2$.
Finally, a weighted average is used to combine the scores of each predicted cluster, with each cluster contributing proportional to its size.
The weighted average score is defined as
\begin{equation}
    \hat{F}^*_w(\cC_1, \cC_2) = \frac{1}{\sum_{C \in \cC_1} |C|} \sum_{C_i \in \cC_1} \left(|C_i| \times \max_{C_j \in \cC_2} F^*(C_i, C_j)\} \right).
\end{equation}
The hat denotes that we are matching communities from $\cC_1$ to communities in $\cC_2$.
Additionally, to consider outliers, we use another weighted average of the cluster similarity and the outlier similarity.
Let $O^{(1)}$ be the set of outliers (nodes not in any cluster) in $\cC_1$, and similarly let $O^{(2)}$ be the set of outliers in $\cC_2$.
Define the outlier aware one-sided similarity as:
\begin{equation}
    \hat{F}^*_{wo}(\cC_1, \cC_2) = \frac{|O^{(1)}|}{|V|}F^*(O^{(1)}, O^{(2)}) + \frac{|V| - |O^{(1)}|}{|V|} \hat{F}^*_w(\cC_1, \cC_2).
\end{equation}
For a symmetric comparison measure, we average the scores from each direction;
\begin{equation}
    F^*_{wo}(\cC_1, \cC_2) = 0.5\hat{F}^*_{wo}(\cC_1, \cC_2) + 0.5\hat{F}^*_{wo}(\cC_2, \cC_1).
\end{equation}

\begin{figure}[t]
    \center
    \includegraphics[width=0.95\linewidth]{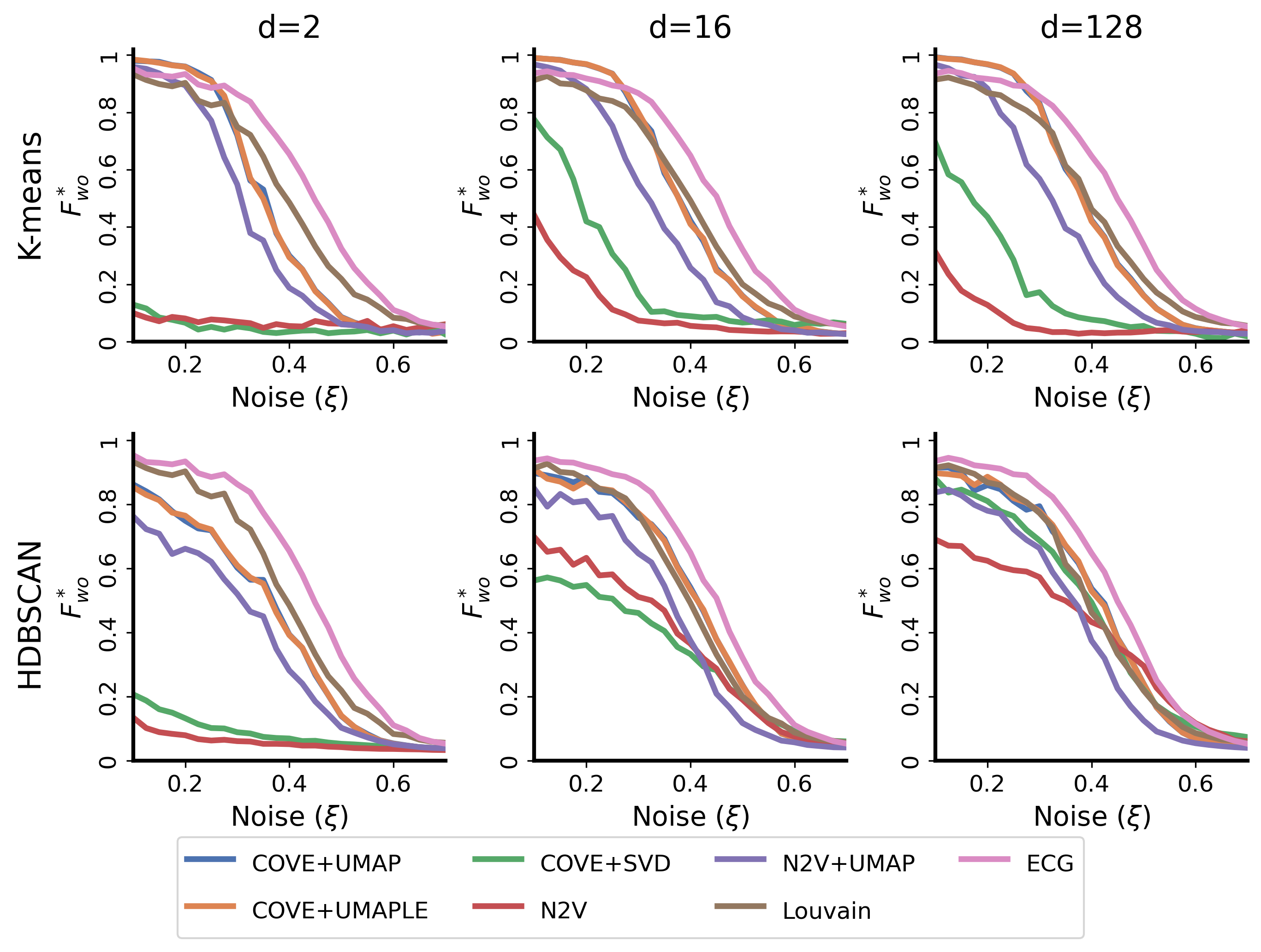}
    \caption{
        Comparing the performance of embedding algorithms for community detection on synthetic ABCD graphs.
        The embeddings are clustered in 2 (left), 16 (center), and 128 (right) dimensions using HDBSCAN (bottom) and K-Means (top).
        The COVE+UMAP and COVE+UMAPLE scores are extremely similar (and difficult to visually distinguish).
        }
    \label{fig:clustering_synthetic}
\end{figure}

In Figure \ref{fig:clustering_synthetic}, we evaluate the embedding methods on a set ABCD~\cite{abcd} graphs.
Each graph has $5000$ nodes and the noise level $\xi$ is varied from $0.1$ to $0.7$ (other parameters are fixed as mentioned in Section \ref{sec:data}).
We generate $10$ samples at each noise level, and report the average $F^*_{wo}$ score between the detected clusters and the ground truth communities.
We fix the minimum cluster size of HDBSCAN left as the default $15$ for all runs, and the number of clusters for K-means is set to the number of ground truth clusters in the target ABCD graph.
For comparison, we also run two modularity based methods directly on the graph: the very popular Louvain~\cite{louvain} algorithm and the state-of-the-art ECG extension~\cite{ecg}.

Across all dimensions (2, 16, or 128), the methods using UMAP or UMAPLE significantly outperform either only node2vec or COVE+SVD.
Furthermore, COVE+UMAP and COVE+UMAPLE perform almost identically, and consistently better than node2vec or node2vec+UMAP.
The K-means clustering method performs well for very low noise $(\xi < \approx 0.3)$, even better than ECG.
In the medium noise range $(0.3 < \xi < 0.5)$, HDBSCAN performs slightly better that K-means, and is comparable to Louvain in dimensions 16 and 128.
Beyond $\xi = 0.5$, none of the methods, including the modularity based options, are not finding any ground truth clusters.

We also test the embedding methods with HDBSCAN on the real graph with ground truth communities.
Since the scale of the ground truth communities varies considerably, we take maximum over a range of minimum cluster sizes (see Appendix \ref{app:min_cluster_size} for more details).
The conclusions are similar to the synthetic tests: COVE+UMAP, COVE+UMAPLE, perform almost identically, and similar to or slightly better than node2vec+UMAP.
In several graphs (most notably primary1, eu-core), the best embedding methods perform better than ECG, although we cannot say it is better overall since we optimized over the minimum cluster size for HDBSCAN, but not over the analogous resolution parameter of modularity, and would like to make clear that Louvain and ECG are included only a reference point for popular methods.
\begin{figure}
    \center
    \includegraphics[width=\linewidth]{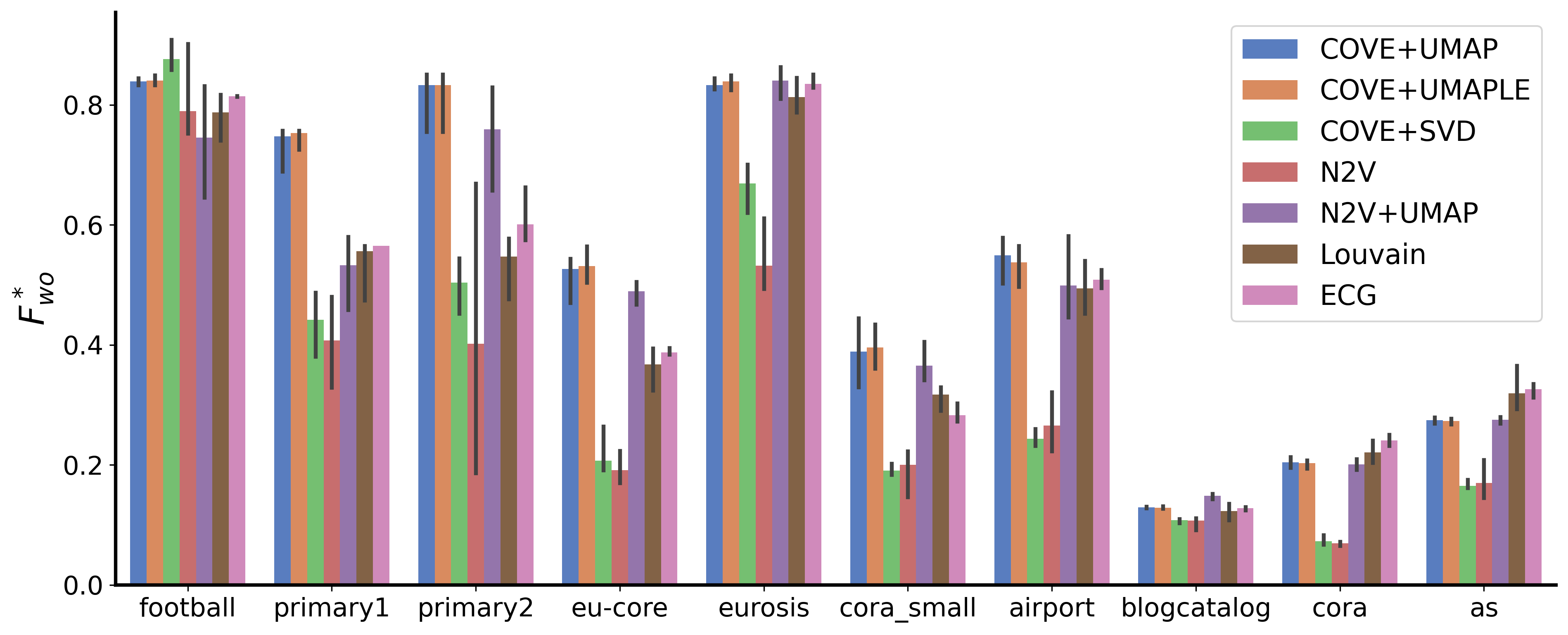}
    \caption{
        Comparing embedding methods for community detection on real graphs with known ground truth communities.
        The embeddings are clustered using HDBSCAN and optimized over the minimum community size parameter.
        Each bar represent the average score over 10 independent embeddings, with the smaller bar spanning the full range of scores.
        COVE+UMAP, COVE+UMAPLE, node2vev+UMAP, Louvain, and ECG perfom similarly and better than COVE+SVD or node2vec on most of the graphs.
        }
    \label{fig:clustering_real}
\end{figure}

\subsection{Link Prediction}
In this experiment, we use the embeddings to train a classifier for predicting missing links.
For each pair of embedded nodes $u$ and $v$, we construct an edge vector $e_{uv}$ using the hadamard product (element-wise multiplication) which has been shown to perform well~\cite{node2vec}.
For each graph, we remove $5\%$ of the edges to use as a test-set, and compute the edge vectors of the remaining edges and an equal number of randomly sampled non-edges (possibly including pairs that are edges in the test set) to use as training data.
We then train a logistic regression classifier and test it on the reserved edges and an equal number of randomly sampled non-edges.

\begin{figure}
    \center
    \includegraphics[width=\linewidth]{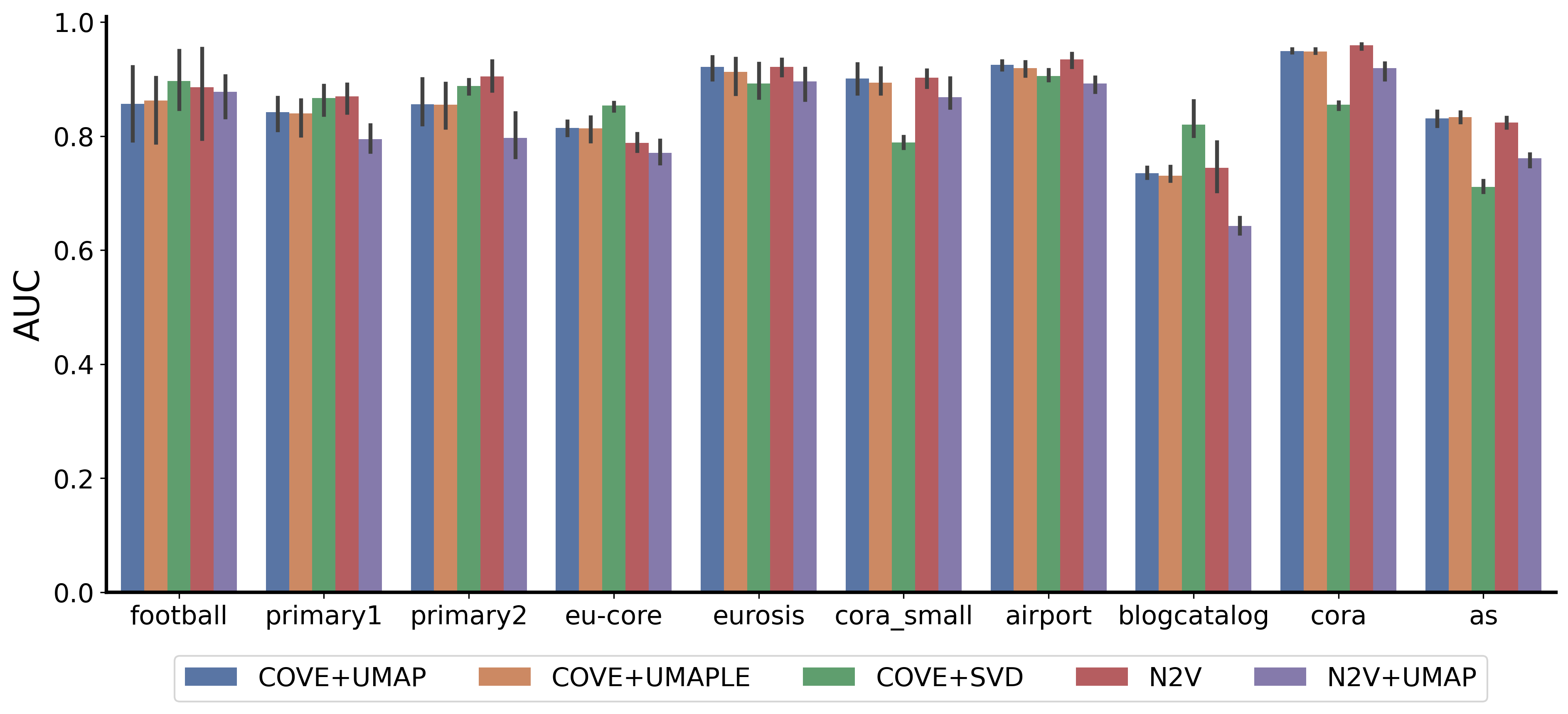}
    \caption{
        Performance of a classifier for link prediction on real graphs using edge vectors created by the hadamard product of the end node embeddings.
        For 10 samples, $5\%$ of edges were removed at random to create a test set and added to an equal number of non-edges to form the test set.
        Each bar represents the average AUC score, with the smaller black bar spanning the full range of values.
    }
    \label{fig:linkpred}
\end{figure}

For each of the real graphs, we repeat the experiment 10 times and report the average performance of the classifier in Figure \ref{fig:linkpred}.
We see very little difference in the performance of each algorithm, and only small variations between runs on the same graph.

\section{Discussion}
In this paper, we propose principled high dimension node embeddings based on co-occurrence in random walks, and leverage modern dimension reduction techniques to achieve similar or slightly improved performance compared to neural embedding methods. 
While an exact computation would be intractable for large graphs, we employ the sampling method used by DeepWalk and node2vec to approximate the embeddings in time that grows linearly with the number of nodes in the graph.
Furthermore, we extended the experiments using embeddings for community detection~\cite{clustering_via_embeddings,neural_embeddings} by using HDBSCAN for clustering, and found that the performance is similar to Louvain on synthetic and real graphs when the embedding is of a moderate dimension.
An interesting direction for future research is UMAP's ability to project to non-euclidean spaces, specifically hyperbolic spaces that have been used in the network science literature~\cite{hyperbolic_embedding,hyperbolic_clustering,hyperbolic_linkpred}, although the interpretation or extension of clustering or link prediction methods to hyperbolic UMAP embeddings would require careful consideration.

\section*{Code Availability}
Code for the method and experiments is available at \href{https://github.com/ryandewolfe33/COVE}{https://github.com/ryandewolfe33/COVE}.
Datasets are publicly available for download from the projects cited in the URL column of Table \ref{ref:tab_data}.

\section*{Acknowledgements}
R.D.~acknowledges the support of the Natural Sciences and Engineering Research Council of Canada (NSERC) via a CGS-M scholarship.

\bibliographystyle{plainurl}
\bibliography{references}

\appendix

\section{Evaluating K-means clustering with the Adjusted Mutual Information}
\begin{figure}[b]
    \center
    \includegraphics[width=\linewidth]{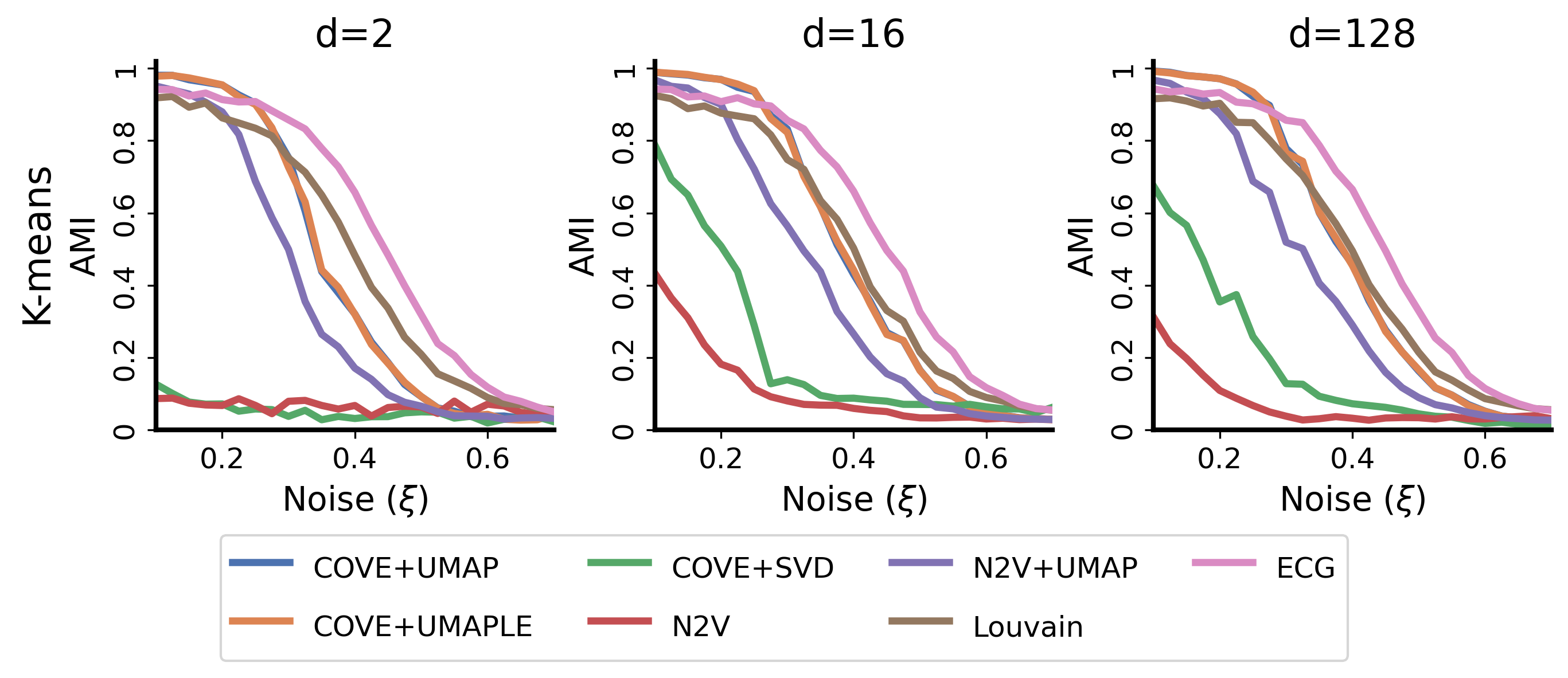}
    \caption{
        Repeating the experiment from Figure \ref{fig:clustering_synthetic}, except using Adjusted Mutual Information to compare clusterings.
        Since the AMI is only defined for partitions, we cannot evaluate results from HDBSCAN.
    }
    \label{fig:ami}
\end{figure}

The main benefit of the $F^*_{wo}$ \cite{f*} score used for clustering comparison in Section \ref{sec:clu} is that it is defined when there are outliers (or overlaps), thus allowing direct comparison between K-means and HDBSCAN.
However, since $F^*_{wo}$ was only recently proposed, we can also use a more familiar measure to evaluate K-Means.
In Figure \ref{fig:ami} we redo the experiment from Figure \ref{fig:clustering_synthetic} using the Adjusted Mutual Information (AMI) \cite{ami} to compare results.
The conclusions are identical to those in Section \ref{sec:clu}: the methods using UMAP perform better than those without, the COVE methods perform slightly better than node2vec, and in moderate or high dimension COVE+UMAPLE+KMeans performs similar to Louvain but still worse than ECG.

\section{Details of Minimum Cluster Size Optimization}\label{app:min_cluster_size}
To account for the different scales of clusters in the real world graphs, and to make sure we are evaluating the embeddings and not the parameter selection of the HDBSCAN, we optimized over the minimum cluster size parameter in Figure \ref{fig:clustering_real}. In particular, for each graph we evaluate each value between $2$ and $15\log_2 |V|$. While it's possible that the optimal value is above this range, most of the optimal values returned are significantly below the maximum value tested as seen in Table \ref{tab:opt}

\begin{table}[h]
    \centering
    \begin{tabular}{|Sc|Sc|Sc|Sc|Sc|}
        \hline
        Graph & Average & Min & Max & $15\log_2 |V|$  \\
        \hline
        Football & 3.36 & 2 & 7 & 102 \\
        primary1 & 5.32 & 2 & 13 & 118 \\
        primary2 & 5.88 & 2 & 13 & 118 \\
        eu-core & 5.00 & 2 & 20 & 149 \\
        eurosis & 13.36 & 4 & 36 & 154 \\
        cora\_small & 30.54 & 4 & 104 & 171 \\
        airport & 64.26 & 6 & 162 & 172 \\
        blogcatalog & 14.86 & 2 & 125 & 199 \\
        cora & 56.42 & 6 & 120 & 217 \\
        as & 70.56 & 10 & 165 & 218 \\
        \hline
    \end{tabular}
    \vspace{1em}
    \caption{Optimal minimum cluster size values for each of the embedding methods and each real graph used in Figure \ref{fig:clustering_real}. Importantly we can see the largest checked value of $15\log_2 |V|$ is larger than the largest optimal value for each graph.}
    \label{tab:opt}
\end{table}

\end{document}